\documentclass[conference]{IEEEtran}
\IEEEoverridecommandlockouts

\usepackage{cite}
\usepackage{amsmath,amssymb,amsfonts}
\usepackage{algorithmic}
\usepackage{graphicx}
\usepackage{textcomp}
\usepackage{xcolor}
\usepackage{colortbl}
\usepackage{multirow}
\usepackage{enumitem}
\usepackage[most]{tcolorbox}
\usepackage{hyperref}
\usepackage{booktabs}
\usepackage{graphicx}
\usepackage{pifont}   
\usepackage{tikz}
\usepackage{array}
\usepackage{booktabs}
\usetikzlibrary{matrix, positioning}
\usepackage{placeins}
\usepackage{url}
\def\BibTeX{{\rm B\kern-.05em{\sc i\kern-.025em b}\kern-.08em
    T\kern-.1667em\lower.7ex\hbox{E}\kern-.125emX}}
\begin{document}


\title{OpsLLM: Construction of Large Language Model for Software Operations with Multi-stage Learning\\
}


\author{

\IEEEauthorblockN{
Jingkai He
}
\IEEEauthorblockA{
School of Systems Science and Engineering\\
Sun Yat-sen University\\
Guangzhou, China\\
hejk25@mail2.sysu.edu.cn
}

\and

\IEEEauthorblockN{
Pengfei Chen$^{*}$
}
\IEEEauthorblockA{
School of Computer Science and Engineering\\
Sun Yat-sen University\\
Guangzhou, China\\
chenpf7@mail.sysu.edu.cn
}

\and
\IEEEauthorblockN{
Chenghui Wu, Shuang Liang
}
\IEEEauthorblockA{
School of Computer Science and Engineering\\
Sun Yat-sen University\\
Guangzhou, China\\
\{wuchh55,liangsh68\}@mail2.sysu.edu.cn
}
\and

\IEEEauthorblockN{
Ye Li
}
\IEEEauthorblockA{
Alibaba Cloud Computing\\
Hangzhou, China\\
liye.li@alibaba-inc.com
}
\and
\IEEEauthorblockN{
Gou Tan
}
\IEEEauthorblockA{
School of Systems Science and Engineering\\
Sun Yat-sen University\\
Guangzhou, China\\
tang29@mail2.sysu.edu.cn
}
\and
\IEEEauthorblockN{
Xidao Wen
}
\IEEEauthorblockA{
Alibaba Cloud Computing\\
Hangzhou, China\\
wenxidao.wxd@alibaba-inc.com
}
\and
\IEEEauthorblockN{
Chuanfu Zhang
}
\IEEEauthorblockA{
School of Systems Science and Engineering\\
Sun Yat-sen University\\
Guangzhou, China\\
zhangchf9@mail.sysu.edu.cn
}
\and

\IEEEauthorblockN{
Fang Situ, Qi Zhou
}
\IEEEauthorblockA{
Alibaba Cloud Computing\\
Hangzhou, China\\
\{jifeng,jackson.zhouq\}@alibaba-inc.com
}

}

\IEEEaftertitletext{\vspace{-1.2cm}}

\maketitle

\begin{abstract}
In the field of software operations, Large Language Models (LLMs) have attracted increasing attention. However, existing research has not yet achieved efficient and effective end-to-end intelligent operations due to low-quality data, fragmented knowledge and insufficient learning. 
  To explore the potential of LLMs in software operations, we propose \textit{OpsLLM}, a domain-specific LLM that supports both knowledge-based question answering (QA) and root cause analysis (RCA). Moreover, we disclose the detailed workflow for building LLMs specifically in the software operations domain. First, a Human-in-the-Loop mechanism is introduced to curate high-quality data from a large collection of operational data and construct a fine-tuning dataset. Then, based on the data, supervised fine-tuning is conducted to achieve a base model. Furthermore,  we introduce a domain process reward model (DPRM) during the reinforcement learning stage to optimize the accuracy and reliability of the fine-tuned model on RCA tasks. 
  Experimental results on the tasks with diverse difficulties demonstrate that \textit{OpsLLMs} effectively learns and aligns with the operational domain knowledge infused, outperforming existing open-source and closed-source LLMs in accuracy with  improvements of \textbf{0.2\%\textasciitilde 11.9\%} on QA tasks and \textbf{8.5\% \textasciitilde  70.3\%} on RCA tasks,  
  while exhibiting strong transferability. Moreover, we will open-source three versions of \textit{OpsLLM} with 7B, 14B and 32B parameters, along with a 15K fine-tuning dataset. 
\end{abstract}


\section{Introduction}


Modern cloud native software systems are rapidly growing in scale and complexity, making software operations increasingly critical. In practice, system failures occur frequently and cause serious incidents. For example, the 2025 Cloudflare outage affected ChatGPT, X, Spotify, and hundreds of thousands of dependent services~\cite{25Cloudflareoutage}. Traditional operational approaches rely on domain-specific rules and fragmented expert knowledge, which generalize poorly across different systems and scenarios. In contrast, Large Language Models (LLMs), with their strong capabilities in knowledge generalization and reasoning, offer a promising direction for intelligent software operations.

Recent studies have applied LLMs to software operations through fine-tuning~\cite{intro1,intro2} or prompt engineering (in-context learning and chain-of-thought prompting)~\cite{intro3,intro4,intro5}. These studies have explored specific tasks such as log parsing~\cite{parser1,parser2}, dialogue generation~\cite{dialogue1,dialogue2}, and code generation~\cite{code1,code3}. However, two key limitations remain: (i) these tasks are addressed in isolation, and core operational capabilities such as domain knowledge QA and RCA are not fully supported; (ii) operational knowledge is highly fragmented, and high-quality labeled data remains scarce.

To achieve effective end-to-end intelligent software operations, models are necessary to support multimodal operational data (e.g., metrics, logs, and traces) and diverse tasks (e.g., QA, configuration analysis, and RCA). We focus on elementary operational tasks represented by QA and advanced operational tasks represented by RCA. It should be noted that we focus on operation problems in cloud-native microservice systems, where Online Boutique and Train Ticket are both representative cloud-native systems. However, building such an LLM faces the following challenges.

\textbf{C1: Cross-modal knowledge alignment under structural heterogeneity.}
Real-world operational knowledge is dispersed across various multimodal operational data sources and textual sources (i.e., web data, open-source communities, software documents, and public datasets). As a result, there is no unified, high-quality labeled dataset for software operations. Furthermore, the large scale of such multimodal data makes manual checking impractical, posing challenges to data quality control. In addition, due to different data structures, it is difficult to use a unified pipeline to transform multimodal data into high-quality QA pairs.

\textbf{C2: Reasoning inconsistency in multi-step root cause analysis.} Existing LLMs achieve a low accuracy on RCA tasks and often show inconsistent reasoning~\cite{c2-1,c2-3}, where the model reaches a correct answer through wrong steps. This problem arises from three aspects. First, RCA involves multi-step reasoning over complex system behaviors, but supervised fine-tuning (SFT) only optimizes final outcomes and cannot ensure correct intermediate steps. Second, reinforcement learning (RL) can provide richer training signals, but lacks reliable process-level evaluation criteria for RCA tasks. Third, jointly optimizing outcomes and reasoning processes easily leads to reward hacking, where models produce seemingly reasonable reasoning that fails to find true root causes.

\textbf{C3: Evaluation ambiguity in operation QA.} LLMs must have accurate operational knowledge to support end-to-end intelligent operations in real-world scenarios. However, existing benchmarks often use plain text as answers, making it hard to judge correctness. The model and ground truth may express the same meaning in different words. As a result, evaluation methods based on textual or semantic similarity struggle to reflect the model's true capability. For example, if the model answers ``restart the service'' and the ground truth is ``reboot the application,'' the similarity score will be low even though both are correct.

\begin{figure*}[t]
    \centering
    \includegraphics[width=0.95\textwidth]{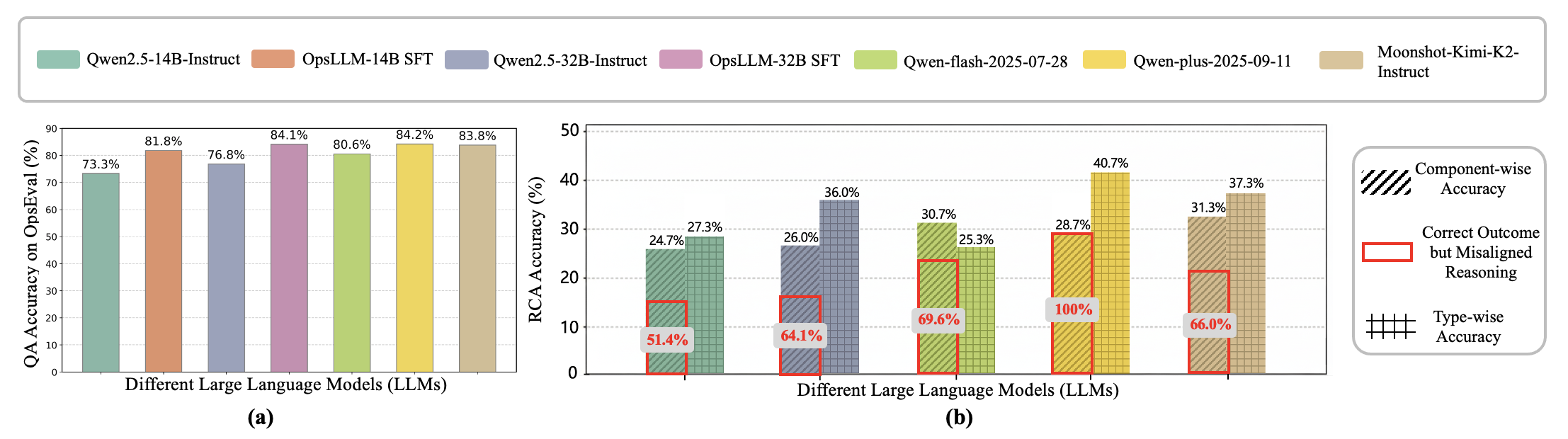}
    \vspace{-0.15in}
    \caption{(a) shows the accuracy of different LLMs on OpsEval. (b) shows the RCA accuracy of different LLMs on 150 fault injection cases from OnlineBoutique. OpsLLM-14B SFT and OpsLLM-32B SFT are fine-tuned models based on Qwen2.5 models.}
    \label{motiv12}
\end{figure*}

\textbf{Operational Reasoning and Alignment.} Existing AIOps studies mainly improve RCA at inference time through prompting, agents, or tool use (e.g., RCACopilot~\cite{rcacopilot}, RCAgent~\cite{rcagent}, and Flow-of-Action~\cite{flow}), while RL-based approaches such as HRLHF~\cite{HRLHF} and ThinkFL~\cite{thinkfl} focus on specific localization objectives. In contrast, we introduce a reproducible post-training workflow for software operations that turns multimodal observability data into an experimentally validated data pipeline, then continuously aligns LLMs with evolving operational knowledge and software-operations reasoning through LoRA ~\cite{lora}, curriculum learning~\cite{curriculum}, GRPO~\cite{grpo} and a process reward model. 

Building upon the conventional post-training paradigm that performs SFT followed by RL, our approach introduces an automated pipeline for instruction tuning data generation and quality filtering over operational data, ensuring that the model acquires high-quality domain knowledge. To address the challenges of insufficient  learning signals caused by the difficulty of exploring correct reasoning trajectories, as well as the misalignment of domain-specific reasoning capabilities in complex operational reasoning tasks, we design a multi-dimensional reward mechanism and develop a domain process reward model to guide the model toward reasoning patterns consistent with domain specific operational  knowledge. Thereby, the model has a strong diagnostic capability for complex RCA tasks.

\textbf{\textit{OpsLLM} Framework.} To address the existing challenges, we propose \textit{OpsLLM} \footnote{Source code: \url{https://github.com/IntelligentDDS/OpsLLM}.}, a domain-specialized LLM for software operations with core capabilities in QA and RCA. 
We design modality-specific data pipelines to unify fragmented operational knowledge into QA pairs and introduce a human-in-the-loop (HITL)~\cite{hitl} mechanism for automated quality control, enabling scalable data construction (\textbf{solution to C1}).
We propose a multi-stage learning framework with a domain process reward model (DPRM) based on GRPO~\cite{grpo} to provide process-level supervision. Combined with curriculum learning~\cite{curriculum} and strict gating, this ensures both accurate RCA outcomes and reliable reasoning (\textbf{solution to C2}).
We also design a multi-expert benchmark generation method for objective evaluation of QA capability (\textbf{solution to C3}). The experimental results indicate that the LLM consistently improves performance, with gains of \textbf{0.2\%\textasciitilde 11.9\%} for QA tasks and \textbf{8.5\%\textasciitilde70.3\%} for RCA tasks.

Generally, we make the following contributions.

\begin{itemize}[]
    \item We propose an end-to-end workflow for building domain-specific LLMs in software operations. It includes: (i) modality-specific data pipelines with a human-in-the-loop mechanism for scalable quality control; (ii) multi-stage post-training combining supervised fine-tuning and reinforcement learning; (iii) a multi-level evaluation benchmark for core operational tasks.
    \item We first propose a domain process reward model (DPRM) for RCA in software operations, based on GRPO to provide process-level supervision. Combined with curriculum learning and strict gating, DPRM ensures both accurate final answers and reliable intermediate reasoning steps.
    \item We evaluate the effectiveness and efficiency of \textbf{\textit{OpsLLM}} at three scales (7B, 14B, and 32B). Moreover, these models and the fine-tuning dataset will be released as open source. Our code including the prompts used in both implementation and experiments, the filtering rules used for dataset construction, the DPRM training data will also be released, as well as the data used in the experiments.
\end{itemize}

\section{Background and Motivation}

\subsection{Background}
In this work, we leverage the observability data in software systems generated by fault injection to conduct QA and RCA tasks. To keep self-contained, we give a brief background. \textbf{Observability data} refers to the data used to monitor the internal status of software systems, typically including metrics (i.e., time-series measurements such as CPU utilization and request latency), traces (i.e., end-to-end request  execution paths across system components), and logs (i.e., unstructured or semi-structured textual records). \textbf{Fault injection} is an important technique for simulating system behavior under abnormal conditions, widely adopted in prior work ~\cite{Fuzztruction,CAFault,AIOpsLab,GAIA-DataSet} on reliability and root cause analysis, which involves deliberately introducing controlled faults (e.g., CPU exhaustion, memory exhaustion, network packet loss) to simulate common failure scenarios in distributed and microservice-based systems for studying failure behavior. \textbf{Root Cause Analysis (RCA)} focuses on identifying the fundamental reasons behind system failures (e.g., performance degradation, service interruptions); when an incident occurs, engineers analyze observability data and system context to determine the root cause—described by \textit{originating component}, \textit{start time}, and \textit{failure reason}—and failures may propagate across components due to dependencies or shared environments, complicating RCA.

\subsection{Motivations}

\textbf{Motivation 1:} Fine-tuning on high-quality operational data is critical to improve LLM operation capabilities, yet constructing such data at scale remains an open challenge under data sensitivity constraints. Operational data are highly correlated with the internal states of private software systems, which cannot be sent to external APIs for certain enterprises. So locally deployed LLMs are necessary. Building such LLMs requires high-quality datasets, yet operational knowledge is fragmented across multiple sources and continuously evolves, making scalable data pipelines essential. Fig.~\ref{motiv12}(a) shows the improvements on QA of fine-tuning existing Qwen2.5-14B/32B-Instruct models (details in Section~\ref{results}).

Building such data pipelines faces two challenges: (i) \textbf{\textit{Lack of unified support for heterogeneous data modalities.}} Operational data spans across metrics, logs, and traces, each requiring different processing logic. (ii) \textbf{\textit{Difficulty in ensuring data quality at scale.}} Operational data grows rapidly and cannot be manually filtered, and low-quality data introduces noise that degrades model accuracy. Therefore, we aim to build scalable data pipelines with automated quality control for locally deployed LLMs.

\textbf{Motivation 2:} Existing LLMs show a low RCA accuracy and unreliable reasoning, requiring learning mechanisms with process-level supervision. in real-world systems, the  RCA procedure directly supports failure localization and remediation. However, existing LLMs achieve low accuracy on RCA tasks. Moreover, some correct answers are achieved with incorrect reasons.

Fig.~\ref{motiv12}(b) shows that Qwen-flash, Qwen-plus, and Kimi-K2-Instruct achieve below 42\% RCA accuracy. Moreover, their reasoning misalignment rates reach 69.6\%, 100\%, and 66.0\%, respectively. Although the root cause component is correctly identified, the incorrect fault level indicates unreliable reasoning.

Improving RCA through SFT alone is limited, since SFT only optimizes final outcomes without constraining intermediate reasoning steps. RL can provide richer supervision, yet faces two challenges: (i) \textbf{\textit{Lack of reliable process-level evaluators.}} RCA involves multi-step reasoning, but supervision is limited to final labels, with few annotated reasoning processes available. (ii) \textbf{\textit{Prone to reward hacking.}} Joint optimization of outcomes and reasoning easily leads to reward hacking, where models produce high-scoring reasoning without improving RCA capability. Therefore, we aim to design a learning framework with process-level supervision to improve both RCA accuracy and reasoning reliability.

\begin{figure}[htbp]
    \centering
    \includegraphics[width=0.9\columnwidth]{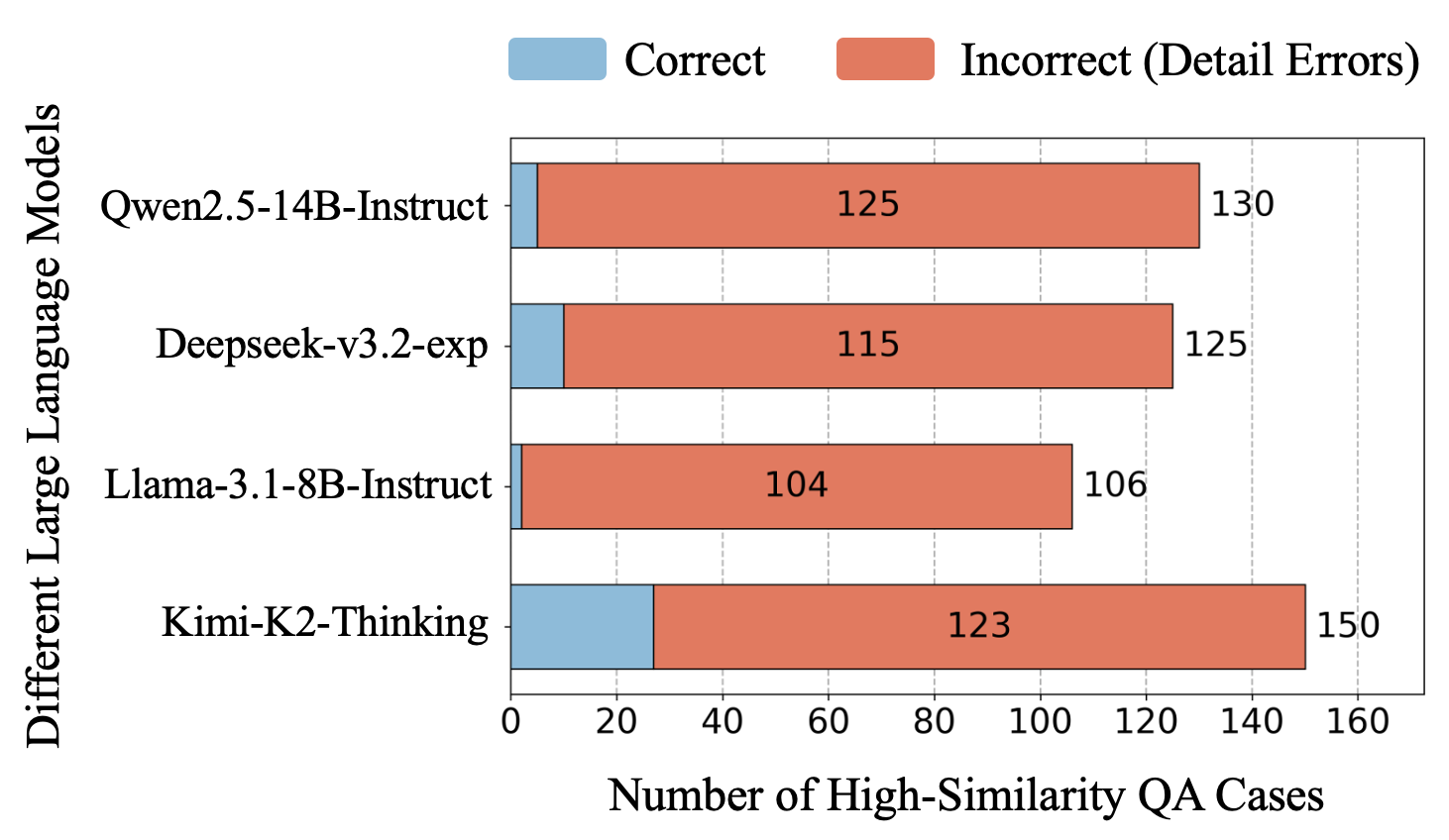}
    \vspace{-0.15in}
    \caption{
    Across different LLMs, more than 80\% of responses ranked in the top 20\% by ROUGE-L/BERTScore still contain detail errors. This indicates that high similarity scores do not guarantee correctness.
    }
    \label{motiv3}
\end{figure}

\begin{figure*}[t]
    \centering
    \includegraphics[width=0.9\textwidth]{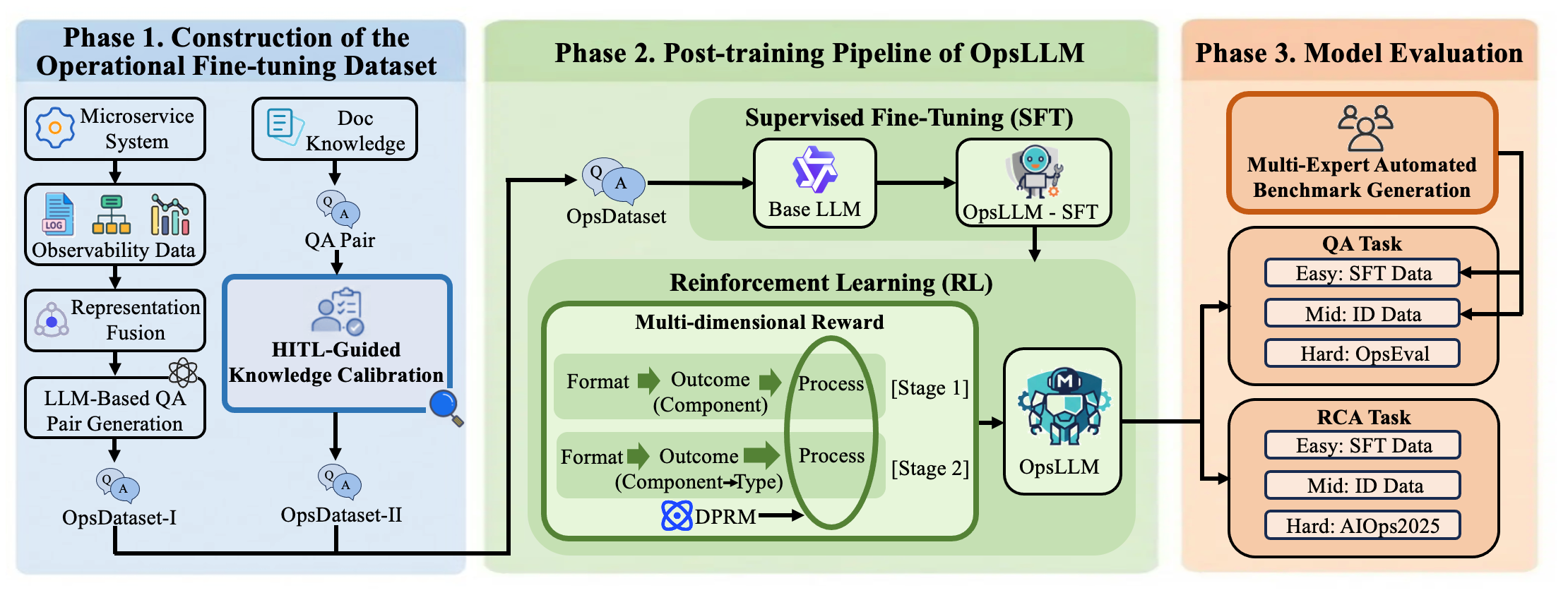}
    \vspace{-0.15in}
    \caption{The workflow of model construction and evaluation.}
    \label{fig:overall}
\end{figure*}

\textbf{Motivation 3:} 
Operational knowledge QA reflects the model's understanding of domain knowledge. However, the similarity-based evaluation is unreliable and inaccurate because models and ground truths often express the same meaning with different words.

To verify this point, we evaluate using two common metrics: ROUGE-L, which measures word-level overlap based on the longest common subsequence, and BERTScore, which measures semantic similarity using pretrained embeddings. In Fig.~\ref{motiv3}, we evaluate different LLM architectures using 1,500 QA instances with the temperature set to 0.2, and the number of cases for each LLM whose ROUGE-L and BERTScore rank in the top 20\% are recorded (the red bar).
It shows that high similarity scores do not guarantee correctness.
Therefore, we aim to transform QA pairs into multiple-choice format, where the correctness can be judged by exact option matching.

\section{Methodology}

\subsection{Overview}

This paper presents \textit{OpsLLM}, a domain-specialized LLM for software operations. As shown in Fig.~\ref{fig:overall}, the framework contains: data construction (\S\ref{phase1}), post-training (\S\ref{phase2}), and evaluation (\S\ref{phase3}). The core innovation of our work lies in prioritizing high-quality real-world data and model capability optimization to directly enhance the LLM’s own software operations expertise, distinguishing it from approaches that treat the LLM as merely a component within external operations frameworks.

\textbf{Task Definition.} We focus on two core tasks: QA and RCA. For QA, given a question $q$ about operational knowledge (e.g., system configuration, troubleshooting), the model generates an answer $a$. For RCA, given multimodal observability data $\mathbf{x} = (x_{\text{metric}}, x_{\text{log}}, x_{\text{trace}})$ collected during a fault, the model predicts the faulty component and the fault type.

\textbf{Phase 1: Data Construction.} We build a fine-tuning dataset $\mathcal{D}$ from two sources: (1) \textit{Observability Data}: we inject faults into microservice systems, fuse multimodal signals into structured representations, and generate RCA samples with reasoning traces. (2) \textit{Doc Knowledge}: we collect textual operational data from diverse sources, convert them into QA pairs, and filter low-quality samples via HITL-guided calibration.

\textbf{Phase 2: Post-training.} We first apply SFT on $\mathcal{D}$ to inject domain knowledge by minimizing:
\begin{equation}
\mathcal{L}_{\text{SFT}}(\theta) = -\sum_{(x,y)\in\mathcal{D}} \log p_\theta(y \mid x),
\end{equation}
where $\theta$ denotes the model parameters, $x$ is the input (question or anomaly information), and $y$ is the target output (answer or reasoning chain). $p_\theta(y \mid x)$ is the probability that the model assigns to output $y$ given input $x$. Minimizing this loss trains the model to produce correct outputs.

Then, we apply RL to improve RCA reliability by maximizing:

\begin{equation}
\label{eq:rl_objective}
\mathcal{J}_{\text{RL}}(\theta) = \mathbb{E}_{y \sim \pi_{\theta}} \left[ R(y) \right] -  \text{KL}(\pi_{\theta} \| \pi_{\text{SFT}}),
\end{equation}
where $\pi_{\theta}$ is the current policy (i.e., the model being trained), $\mathbb{E}[\cdot]$ denotes expectation over sampled outputs, and $R(y)$ is the reward that integrates format correctness, outcome accuracy, and process quality via a stage-wise gating mechanism (scored by DPRM). The KL measures the divergence between $\pi_{\theta}$ and the SFT model $\pi_{\text{SFT}}$.

\textbf{Phase 3: Evaluation.} We evaluate on QA and RCA tasks at three difficulty levels: easy (fine-tuning data), mid (in-distribution data unseen during training), and hard (external benchmarks). For QA, we design a multi-expert automated benchmark to ensure objective evaluation for questions at the easy and mid difficulty levels.

\begin{figure}[htbp]
    \centering
    \includegraphics[width=\columnwidth]{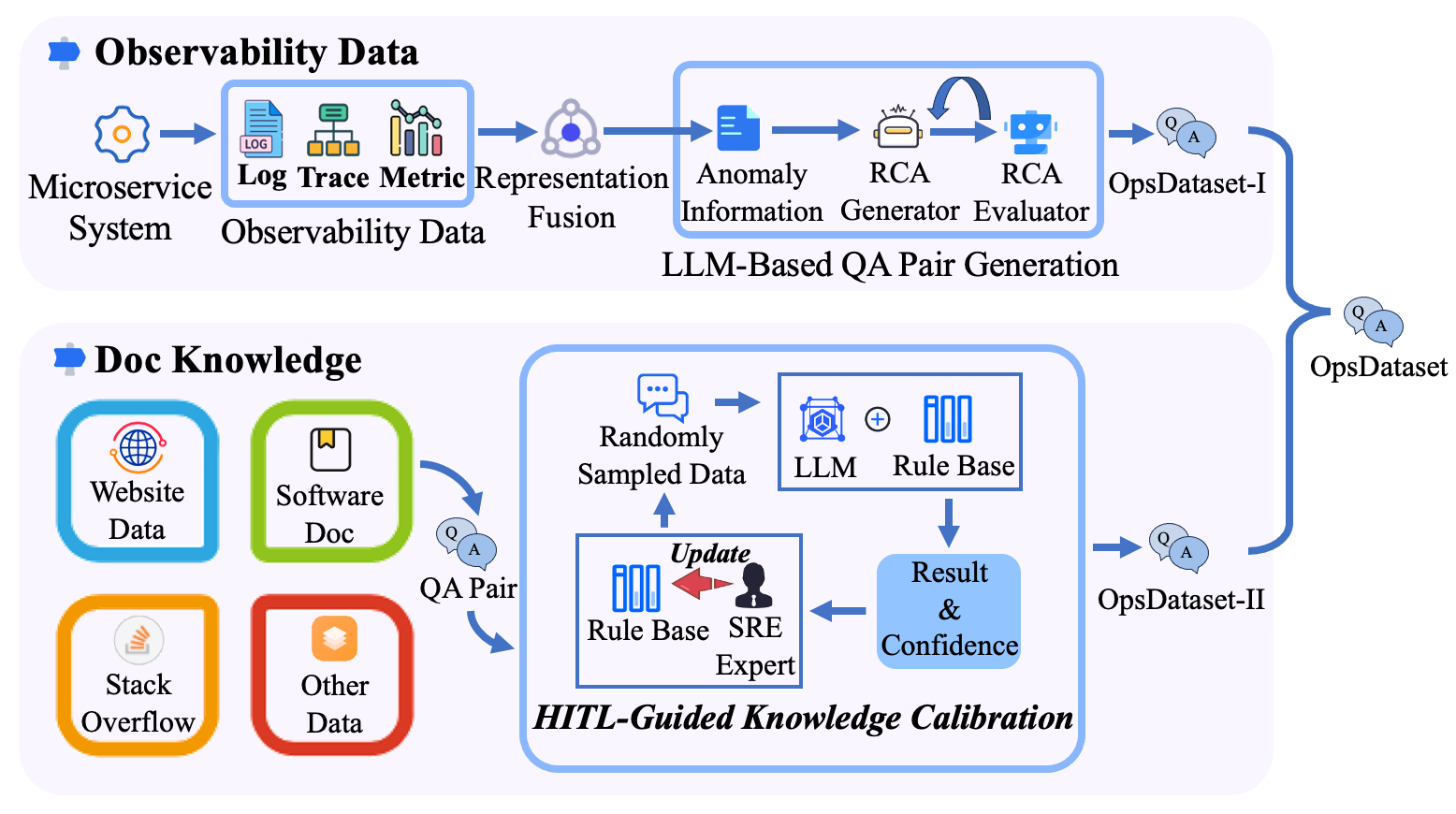}
    \vspace{-0.3in}
    \caption{Ops fine-tuning dataset construction workflow.}
    \label{phase1.png}
\end{figure}

\subsection{Construction of Fine-tuning Operation Data}
\label{phase1}

To address C1 (i.e., fragmented knowledge), we construct two automated pipelines to build high-quality fine-tuning data from large-scale operational data, as shown in Fig.~\ref{phase1.png}. Based on data modality, we process observability data and textual data separately. Both pipelines reduce manual annotation overhead.

\textbf{\textit{Observability Data.}} We generate RCA data from controlled fault-injection experiments on OnlineBoutique\footnote{\url{https://github.com/GoogleCloudPlatform/microservices-demo}}, a Kubernetes-based microservice benchmark widely used in prior RCA studies~\cite{RCAEval, TraFaultDia}. OnlineBoutique provides realistic service interactions, telemetry signals, and failure-propagation behaviors, making it a representative testbed for operational analysis. In total, 596 QA pairs in the fine-tuning dataset are derived from observability data.

\textbf{Representation Fusion.} 
\label{Representation Fusion}
To construct a high-fidelity and reproducible RCA dataset, we implement an automated fault-injection pipeline on the Kubernetes-based Online Boutique system using Chaos Mesh. This enables reproducible chaos engineering experiments within the Kubernetes cluster. For each experiment, a target microservice is randomly selected. Moreover, representative failures, including CPU hog, memory exhaustion, network latency, and packet loss, are injected via StressChaos~\cite{ChaosMeshStressChaos} and NetworkChaos~\cite{ChaosMeshNetworkChaos} for 2–5 minutes with stochastic intensities. During fault injection, metrics, logs, and distributed traces are continuously collected. Meanwhile the fault type, target component, and injection interval are automatically recorded as ground-truth labels.

Next, raw observability data are transformed into compact RCA samples through modality-specific anomaly detection and signal extraction. Metric anomalies are detected using a rolling $3\sigma$ rule within a 1-minute sliding window and summarized by anomaly type, severity, and affected components. Logs are filtered to retain failure-indicative entries (e.g., ERROR, CRITICAL, and FATAL), while traces are extracted if they contain non-zero gRPC or 5xx status codes or exceed the historical P95 latency baseline. These criteria capture both explicit execution failures and performance degradations that are commonly associated with root-cause propagation.

The extracted multimodal signals are organized into RCA samples, each represented as: 
\begin{equation}
s = (\{x^{(m)}\}_{m \in \mathcal{O}}, y),
\end{equation}
where $\mathcal{O}$ denotes the observation space of available modalities, including logs, metrics, and traces. The extracted signals from different modalities are aligned with a fault window and component and fused into a unified structured representation. The resulting multimodal representation is directly used as the query in the LLM-based RCA task. By retaining only information essential for RCA, this process quantitatively achieves a compression ratio of over 99\,\%, condensing approximately 60\,MB of raw telemetry data into a 20\,KB structured prompt while preserving all signals required by the LLM for accurate reasoning.

\textbf{LLM-Based QA Pair Generation.} We augment each query with a step-by-step reasoning trace. As shown in Fig.~\ref{phase1.png}, the RCA Generator (Qwen-Max-2025-09-23) produces an initial trace based on the fused representation. To ensure faithfulness, we employ a reflection-and-verification mechanism: if the deduced root cause mismatches the ground truth, the generator reflects and regenerates the trace. The RCA Evaluator (LLM with different prompt) validates and only retains the traces with logical consistency.

\textbf{\textit{Doc Knowledge.}} Textual operational data are collected from four sources:  \textcircled{1} \textit{Website Data} (1,983 QA pairs) crawled from operations sites; \textcircled{2} \textit{Software Documentation} (5,320 QA pairs) covering OS (including Linux), DB (including MySQL and etcd), Runtime (including JVM, Node.js, and .NET), Framework (including Spring Boot and Flask), and Cloud Infrastructure (including Docker and Kubernetes). \textcircled{3} \textit{Stack Overflow} (4,915 pairs) derived from resolved issues; and \textcircled{4} \textit{Other Data} (2,213 pairs) from six open-source datasets~\cite{owl,devopsllm,AIOps_training,aiops,aiops_camel,aiops_finetuning_qaset}. In total, the document knowledge pipeline produces 14,431 QA pairs.

Crawled data and documentation are standardized into Markdown and converted to QA pairs via Easy DataSet~\cite{easydataset}.

We then employ the following \textbf{human-in-the-loop (HITL)-Guided Knowledge Calibration} module for quality control:

\textbf{Initial Sampling and Annotation.} We construct a 1,000-pair seed set through weighted sampling across sources. Each QA pair is then labeled by expert annotator as \textit{High-Quality} or \textit{Low-Quality} based on initial rules $R_0$. Two experts independently labeled the seed set, resolving disagreements through discussion, achieving high agreement (Cohen’s $\kappa$ = 0.89).The total annotation effort was 20 person-hours (10 hours per expert, in parallel), comprising 8 hours for independent labeling, 1 hour for cross-validation and discussion, and 1 hour for consistency checking and documentation.

\textbf{LLM-based Filtering with Confidence Estimation.} We prompt Qwen-Max-2025-09-23 to evaluate the seed set using $R_0$, outputting a judgment (\textit{Accept/Reject}) and a confidence level (\textit{High/Medium}). The high confidence requires a clear, rule-grounded rationale.

\textbf{Iterative Rule Refinement.} The process begins with an initial rule set $R_0$ that encodes coarse criteria regarding AIOps relevance, technical specificity, and answer detail. We refine $R_k$ to $R_{k+1}$ at the $k$-th iteration by analyzing discrepancies where the LLM either has low confidence or accepts human-labeled low-quality pairs. Through this inspection, the rules are progressively concretized, for instance by prohibiting link-only responses, enforcing linguistic consistency, and disambiguating broad constraints via few-shot counterexamples. This iterative cycle continues until the high-confidence agreement rate reaches $\theta = 0.95$. More precisely, we require
$
\frac{1}{N} \sum_{i=1}^{N} \mathbb{I}(\hat{y}_i = y_i \land c_i = \text{High}) \geq \theta
$, 
where $\hat{y}_i$ and $y_i$ denote the judgments from the LLM and the expert for the $i$-th pair respectively, $c_i$ represents the LLM's confidence, $\mathbb{I}$ is the indicator function. This process yields a final rule set $R_{final}$ that ensures near-perfect agreement between the LLM and experts. The specific contents of $R_{final}$ can be found in the open-source code.

\textbf{Full-Scale Automated Filtering.} Using the final rule set, the LLM automatically filters the entire unlabeled corpus,
aligning with human preferences at a significantly reduced cost. Specifically, we retain only samples predicted as high-quality with high confidence, and discard the medium-confidence ones (about 6\% of the total) to avoid potential noise.

In summary, the 15K fine-tuning dataset $\mathcal{D}$ contains 596 QA pairs from observability data and 14,431 pairs from document knowledge. The fine-tuning dataset is constructed from multimodal operational data sources, but the resulting training instances are uniformly mapped into text-form instruction-supervision samples during training.

\subsection{Post-training Pipeline of \textit{OpsLLM}}
\label{phase2}

\textbf{Supervised Fine-tuning (SFT).} First, using the constructed fine-tuning dataset, we fine-tune Qwen2.5~\cite{qwen2.5} models with LoRA~\cite{lora} based on LlamaFactory~\cite{llamafactory}, denoted as OpsLLM-7B/14B/32B SFT.

\textbf{Reinforcement Learning (RL).} To improve both accuracy and reasoning reliability of LLMs on RCA tasks, we implement RL using Verl~\cite{verl} with GRPO~\cite{grpo} as the optimization algorithm, following prior work~\cite{choosegrpo,grpo1,grpo3}. We use 1,538 RCA samples from fault injection on OnlineBoutique, processed with the representation fusion method (Section~\ref{Representation Fusion}). We address the difficulty of learning complex operational diagnostics, where correct answers may arise from spurious or misaligned reasoning. The reward $R(y)$ used in Eq.~(\ref{eq:rl_objective}) is instantiated as a stage-wise gated reward:

\vspace{-0.05in}
\begin{equation}
R(y)
=
\begin{cases}
g_{1}(y)\, R_{\mathrm{DPRM}}(y),
& \text{if } \mathbb{I}_{\text{stage}=1}=1, \\[4pt]
g_{2}(y)\, R_{\mathrm{DPRM}}(y),
& \text{if } \mathbb{I}_{\text{stage}=2}=1.
\end{cases}
\end{equation}
\vspace{-0.05in}

\noindent
where the gating functions $g_{1}(y), g_{2}(y)\in\{0,1\}$ implement a stage-wise curriculum across two sequential training stages. At Stage~1, $g_{1}$ enforces output format and component correctness before applying the process reward. At Stage~2, $g_{2}$ further requires correct fault-type prediction. $\mathbb{I}_{\text{stage}}$ denotes the indicator of the current training stage, ensuring that the DPRM-based process reward $R_{\mathrm{DPRM}}(y)$ is applied only after the corresponding stage constraints are satisfied. The KL regularization term in Eq.~(\ref{eq:rl_objective}) is applied throughout both stages, while this section focuses on the design of $R(y)$.

The key challenge is designing stable rewards for intermediate diagnosis. To address this, we design the following two components to enable coordinated optimization of outcome correctness and reasoning consistency in RCA tasks.

\textbf{\underline{D}omain \underline{P}rocess \underline{R}eward \underline{M}odel (DPRM).} DPRM is a lightweight model that scores the quality of RCA reasoning chains. We design five RCA-specific scoring rubrics using GPT-5.2~\cite{gpt}, verified by operations experts: \textit{evidence grounding}, \textit{topology consistency}, \textit{causal completeness}, \textit{prediction support}, and \textit{logical consistency}. 

The main contents are as follows: \textcircled{1} \textit{evidence grounding}. This dimension evaluates whether the reasoning is grounded in concrete anomaly evidence provided in the input, such as metrics, timestamps, and affected components. The purpose is to prevent hallucinated or unsupported claims and to ensure that each diagnostic step is anchored in observable system behavior. \textcircled{2} \textit{topology consistency}. This dimension examines whether the reasoning correctly references system topology and respects valid directional relationships between components. It is designed to discourage invalid causal propagation and to ensure that the inferred fault paths are structurally consistent with the actual system architecture. \textcircled{3} \textit{causal completeness}. This dimension assesses whether the reasoning forms a coherent multi-step causal chain that connects the suspected fault to downstream anomalies. The goal is to avoid fragmented or speculative explanations by encouraging causally complete reasoning without unexplained leaps. \textcircled{4} \textit{prediction support}. This dimension measures whether the final root-cause prediction is explicitly supported by the preceding reasoning and anomaly evidence. It aims to align the predicted component and fault type with the diagnostic explanation, reducing cases where the conclusion contradicts or is weakly justified by the reasoning process. \textcircled{5} \textit{logical consistency}. This dimension evaluates the internal consistency of the reasoning, including the absence of contradictory statements and fabricated evidence. Its purpose is to enforce globally coherent explanations and to penalize reasoning patterns that violate basic logical or structural constraints.

To construct DPRM, we use 1,615 fault-injection cases (different from RL training data). For each case, we generate diagnostic reasoning chains with Qwen2.5-3B-Instruct based on anomaly information from multimodal data (Section~\ref{Representation Fusion}), sampling five responses at temperature 0.7. Since DPRM evaluates reasoning quality given a known root cause, we use the correct component and type as input. Each chain is scored by Qwen-Max-2025-09-23 according to the rubrics, assigning one point per satisfied criterion (0--5 scale). The scored chains form training data for fine-tuning Qwen2.5-3B-Instruct with LoRA~\cite{lora} via LlamaFactory~\cite{llamafactory}, yielding a lightweight DPRM. The toal 8,075 samples have the score distributions: 1,670 (0), 2,108 (1), 1,892 (2), 1,192 (3), 705 (4), and 508 (5).

\textbf{Multi-dimensional Reward.} Using DPRM, we adopt a two-stage curriculum training, as shown in Fig.~\ref{fig:overall}. In each stage, rewards follow a gated sequence: each criterion is evaluated only if all preceding ones pass.

\textit{Stage 1} focuses on component localization. Rewards are computed for: (1) Format: output format correctness (1/0); (2) Outcome: diagnostic level (1/0) and component accuracy (1/0), where the component is evaluated only if the level is correct; (3) Process: DPRM score for reasoning quality (0--5). The total reward is the sum of these three terms and is then normalized to the range of 0 to 1. Diagnostic level checks whether the predicted fault granularity matches the ground truth.

\textit{Stage 2} adds fault type prediction. Rewards are computed for: (1) Format (1/0); (2) Outcome: diagnostic level (1/0), component accuracy (1/0), and fault type accuracy (1/0), where the component is evaluated only if the level is correct, and the fault type is evaluated only if the component is correct; (3) Process: DPRM score (0--5), applied only when both component and type are correct. The total reward is computed as the sum of these three terms and then normalized to lie between 0 and 1. Requiring component correctness preserves the localization capability from Stage 1. When the normalized total reward in Stage 1 has stably converged to a level at which the model can obtain process rewards ($>$ 0.375, i.e., $3/8$, where 3 is the score for correct format, level, and component prediction, and 8 is the maximum total reward in Stage 1 after the process reward is introduced) and no longer shows sustained improvement, we consider the model to have stably mastered component localization and switch to Stage 2.

\begin{figure}[t]
    \centering
    \includegraphics[width=0.9\linewidth]{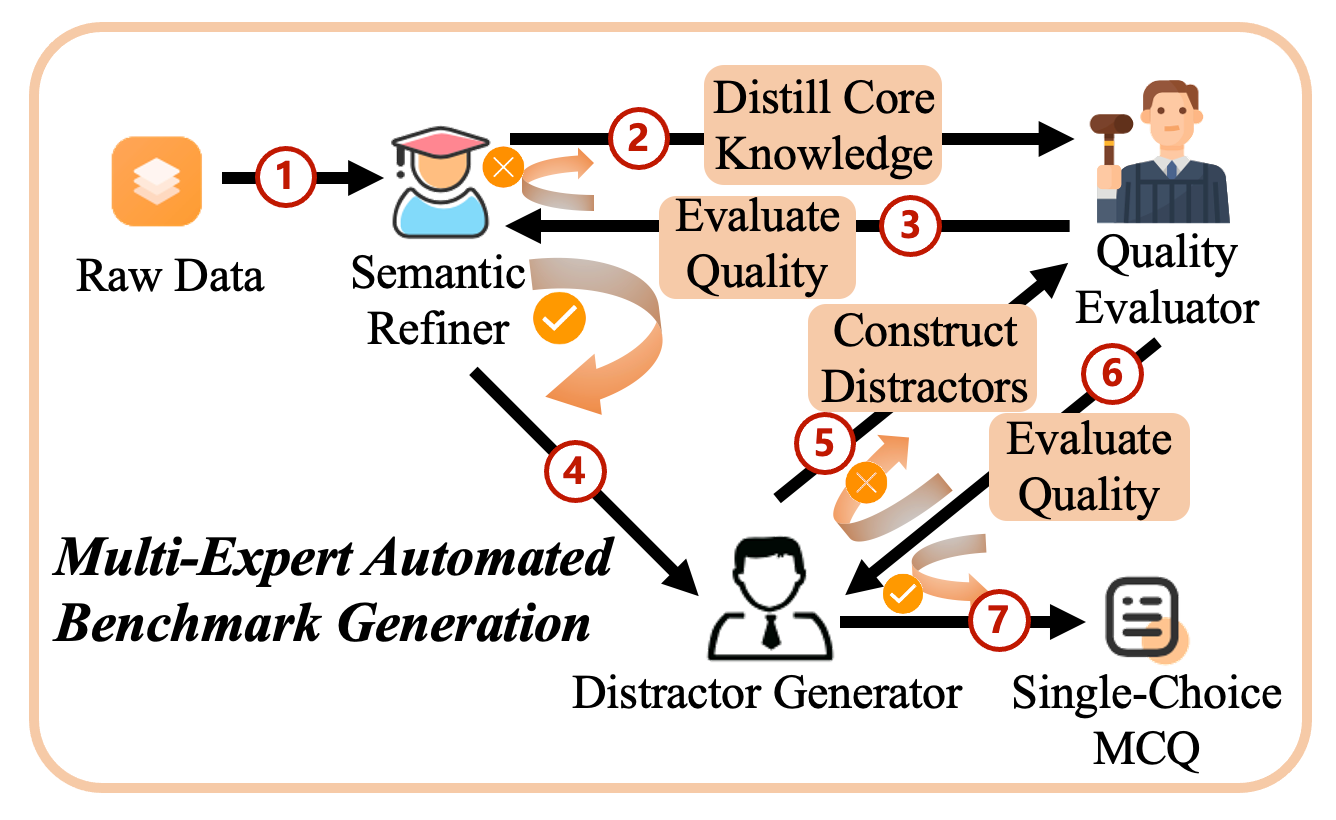}
    \vspace{-0.15in}
    \caption{Workflow of QA pairs to single-choice MCQ.}
    \label{moa.png}
\end{figure}

\subsection{Model Evaluation}
\label{phase3}

We evaluate OpsLLM on two core tasks: QA and RCA. For each task, we design three difficulty levels: (1) \textit{Easy} uses fine-tuning data to test whether the model remembers injected knowledge; (2) \textit{Mid} uses in-distribution (ID) data unseen during training to test generalization within the same domain; (3) \textit{Hard} uses external benchmarks (OpsEval~\cite{opseval} for QA, AIOps2025~\cite{aiops2025} for RCA) to test transferability to new scenarios.

\textbf{QA Evaluation.} To address C3 (lack of objective QA evaluation), we construct a high-quality Multiple Choice Question (MCQ) benchmark using a multi-agent framework shown in Fig.~\ref{moa.png}. This module comprises a Semantic Refiner $\mathcal{R}$, a Distractor Generator $\mathcal{D}$, and a Quality Evaluator $\mathcal{E}$ who engage in iterative refinement.

Formally, let $f^{(k)} = \mathcal{E}(s^{(k)})$ (or $f^{(k)} = \mathcal{E}(m^{(k)})$) represents the critique from the evaluator, $s^{(k)}$ and $m^{(k)}$ denote summary and MCQ in $k$-th iteration. The generation of the optimal summary $s^*$ and MCQ $m^*$ is modeled as two sequential feedback loops:

\begin{equation}
\begin{aligned}
    \text{Loop 1:} \quad & s^{(k)} = \mathcal{R}(s^{(k-1)}, f^{(k-1)}) , k \ge 1 \\
    \text{Loop 2:} \quad & m^{(k)} = \mathcal{D}(m^{(k-1)}, f^{(k-1)}, s^*), k \ge 1
\end{aligned}
\end{equation}

\textit{Loop 1: Knowledge Distillation (Steps \textcircled{1}--\textcircled{3}).} 
$\mathcal{R}$ initializes the process by distilling raw data into a summary $s^{(0)}$, and then $\mathcal{E}$ audits $s^{(0)}$ for hallucinations. If rejected, $\mathcal{E}$ generates feedback $f$, guiding $\mathcal{R}$ to produce a corrected version $s^{(1)}$. This iteration terminates when $\mathcal{E}$ outputs a \textit{Pass} decision, yielding the final $s^*$.

\textit{Loop 2: Distractor Construction (Steps \textcircled{4}--\textcircled{7}).} 
Leveraging $s^*$, $\mathcal{D}$ constructs a candidate MCQ $m^{(0)}$, and then $\mathcal{E}$ rigorously validates the logical exclusivity of the distractors. Similar to Loop 1, any ambiguity triggers feedback-driven regeneration, ensuring the final benchmark sample $m^*$ is factually and logically sound.

\textbf{RCA Evaluation.} For RCA, we use two metrics that align with the two-stage RL training (Section~\ref{phase2}): (1) \textit{Component-level Accuracy} measures whether the predicted faulty component matches the ground truth, corresponding to Stage 1; (2) \textit{Type-level Accuracy} measures whether both the component and fault type are correct, corresponding to Stage 2.

\begin{table*}[htbp]
\caption{Performance comparison across different LLMs.}

\label{tab:llm_acc_comparison}

\resizebox{\textwidth}{!}{
\begin{tabular}{
>{\centering\arraybackslash}m{0.12\textwidth}
>{\centering\arraybackslash}m{0.22\textwidth}
*{6}{>{\centering\arraybackslash}m{0.09\textwidth}}
}
\toprule
\multirow{2}{*}{LLM Category} 
& \multirow{2}{*}{LLM} 
& \multicolumn{6}{c}{ACC (\%)} \\
\cmidrule(lr){3-8}
& 
& \multicolumn{3}{c}{QA} 
& \multicolumn{3}{c}{RCA} \\
& 
& Easy & Mid & Hard 
& Easy & Mid & Hard \\
\midrule

\multirow{3}{*}{Base LLM}
& Qwen2.5-7B-Instruct 
& 96.8 & 97.0 & 66.1 & 8.1 & 4.0 & 4.4 \\
& Qwen2.5-14B-Instruct 
& 97.6 & 97.2 & 73.3 & 9.7 & 4.3 & 4.7 \\

& Qwen2.5-32B-Instruct 
& 98.0 & 98.8 & 76.8 & 11.3 & 8.7 & 6.0 \\
\midrule

\multirow{9}{*}{Open-source LLM}
& Moonshot-Kimi-K2-Instruct
& 98.6 & 98.6 & 83.8 & 11.3 & 11.7 & 7.7 \\
& Deepseek-v3.2-exp
& 96.6 & 97.6 & 85.6 & 9.7 & 10.0 & 19.2 \\
& Qwen3-Next-80B-A3B-Thinking
& 96.2 & 96.8 & 83.8 & 22.6 & 11.7 & 27.4 \\
& aiops-qwen-4b
& 95.0 & 95.8 & 58.7 & 8.1 & 2.3 & 1.6 \\
& Zhiyu2.0-32B 
& 98.4 & 99.2 & 8.6 & 35.5 & 10.0 & 0.3 \\
& R1-Distill-SRE-Qwen-7B
& 40.0 & 38.6 & 35.8 & 3.2 & 0.0 & 0.0 \\
& R1-Distill-SRE-Qwen-32B-INT8
& 95.4 & 96.6 & 66.6 & 12.9 & 16.7 & 6.3 \\
& Qwen3.6-35B-A3B
& 77.2 & 76.6 & 84.1 & 12.9 & 10.7 & 17.3 \\ 
\midrule

\multirow{6}{*}{Closed-source LLM}
& GPT-5.2
& 98.8 & 98.4 & 84.5 & 22.6 & 10.3 & 21.6 \\
& Qwen-Plus-2025-09-11
& 96.0 & 97.8 & 84.2 & 21.0 & 11.7 & 24.7 \\
& Qwen-Turbo-2025-07-15
& 96.4 & 98.2 & 76.0 & 6.5 & 11.0 & 13.4 \\
& Qwen3-Max-2025-09-23
& 96.6 & 99.0 & 85.9 & 21.0 & 10.0 & 22.2     \\
\midrule
\rowcolor{blue!10}
      & OpsLLM-7B 
      & 98.0 & 97.6 & 66.3 & 38.7 & 41.3 & 12.9 \\

\rowcolor{blue!10}
OpsLLM & OpsLLM-14B 
& 99.2 & 98.4 & 79.0 & 53.2 & 56.0 & 21.1 \\
\rowcolor{blue!10}
      & OpsLLM-32B 
      & \textbf{99.6} & \textbf{99.8} & \textbf{88.7} & \textbf{77.4} & \textbf{79.0} & \textbf{32.3} \\

\bottomrule

\end{tabular}

}
\end{table*}

\section{Experimental Evaluation}

We evaluate \textit{OpsLLM} to answer three research questions (RQs):

\begin{itemize}
    \item \textbf{RQ1}: How effective is \textit{OpsLLM} in knowledge-based QA and Root Cause Analysis for software operations?
    \item \textbf{RQ2}: How transferable is \textit{OpsLLM} across different test sets, larger model scales and different microservice systems?
    \item \textbf{RQ3}: What is the contribution of each design module?
\end{itemize}

\subsection{Experimental Settings}

\textbf{Implementation.} We conduct experiments on NVIDIA A100-80GB $\times$ 8 GPUs. We use Python 3.10 and vLLM v0.8.2. For SFT, we use LoRA (rank=64), set the learning rate to $3 \times 10^{-5}$, use a per-device training batch size of 1. For RL, we use LoRA (rank=64, alpha=32), with a learning rate of $3 \times 10^{-6}$, batch size of 32, KL coefficient ($\beta$) of 0.001, 32 generations per prompt.

\textbf{Scope Definition of QA and RCA Tasks.} We focus on QA and RCA because QA covers abundant software operations activities including software installation, software troubleshooting, configuration and et.al., while RCA remains a fundamental challenge where existing LLMs still perform poorly (Section~\ref{rca_fail}). This choice is also consistent with recent software operations LLMs by Tencent~\cite{zhiyu}. Moreover, the fault injection serves as a data generation mechanism by creating realistic scenarios.

\textbf{Datasets.} We evaluate on QA and RCA tasks at three difficulty levels, as shown in Table~\ref{tab:dataset}. 
\textit{Easy} samples are drawn from SFT training data (seen during training), testing whether the model retains learned knowledge. 
\textit{Mid} samples are held-out data not used in any training stage (unseen), testing generalization to new but in-distribution queries. 
\textit{Hard} samples come from external public benchmarks, testing transferability to out-of-distribution scenarios.

For QA, Easy and Mid are converted to MCQs using the method in Section~\ref{phase3}. The hard split adopts OpsEval~\cite{opseval} for transferability evaluation. This dataset is excluded from all stages of model construction and serves as a comprehensive AIOps benchmark that provides domain-specific datasets for systematically assessing LLM performance in software operations. For RCA, Easy and Mid are processed with representation fusion (Section~\ref{Representation Fusion}), and all candidate components and fault types are provided as context. The hard split evaluates transferability using Track 1 data from the AIOps2025 Challenge~\cite{aiops2025}, which is excluded from all stages of model construction and focuses on accurate and efficient root cause localization in complex microservice architectures. Similarly, all candidate components and fault types are provided as context. The CCF AIOps 2025 Track 1 dataset is derived from a Kubernetes-based microservice environment and can be conservatively regarded as representative of cloud-native microservice systems.

\textbf{Baselines.} we compare with three categories of LLMs (Table~\ref{tab:llm_acc_comparison}):
\begin{itemize}
    \item\textit{Base LLMs}: Qwen2.5-7B/14B/32B-Instruct~\cite{qwen2.5}. These show baseline capability before fine-tuning.   \item\textit{Open-source LLMs}: Moonshot-Kimi-K2-Instruct, Deepseek-v3.2-exp, Qwen3-Next-80B-A3B-Thinking, aiops-qwen-4b, Zhiyu2.0-32B, R1-Distill-SRE-Qwen-7B, R1-Distill-SRE-Qwen-32B-INT8 and Qwen3.6-35B-A3B. These are general-purpose LLMs and domain-specific LLMs on operational tasks. 
    \item\textit{Closed-source LLMs}: Qwen-Plus-2025-09-11, Qwen-Turbo-2025-07-15, Qwen3-Max-2025-09-23 and GPT-5.2. These represent commercial LLMs.
\end{itemize}
\vspace{-0.05in}
Note that these baselines span multiple scales and contain both general-purpose and domain-specific models, as well as both open-source and closed-source ones.

\textbf{Metrics.} For QA, we use accuracy (ACC): the proportion of correctly answered MCQs. For RCA, we also use ACC. For each case, the model is restricted to output exactly one root-cause component, one fault type, and the corresponding diagnostic process. A case is considered correct only if the predicted component and fault type exactly match the ground-truth strings. In addition, component diagnosis is considered correct only when the model identifies the correct fault level (including service-level, pod-level, and node-level), with node-level referring to failures caused by a specific node, pod-level referring to failures of an individual container, and service-level referring to cases where all pods of a service are faulty and the service itself constitutes the root cause.

\begin{table}[htbp]
\centering
\caption{Evaluation dataset for different difficulty-levels.}

\label{tab:dataset}
\resizebox{\columnwidth}{!}{%
\begin{tabular}{lccc}
\toprule
\textbf{Task} & \textbf{Easy\textsuperscript{$\dagger$}} & \textbf{Mid\textsuperscript{$\ddagger$}} & \textbf{Hard} \\
\midrule
QA  & SFT data (500) & HITL-filtered (500) & OpsEval~\cite{opseval} (1945) \\
RCA & SFT data (62) & Fault injection (300) & AIOps2025~\cite{aiops2025} (365) \\
\bottomrule
\end{tabular}
}

\raggedright
\footnotesize
\textsuperscript{$\dagger$}Seen: included in SFT training. 
\textsuperscript{$\ddagger$}Unseen: excluded from all training.
[]: Number.
\end{table}

\begin{figure*}[t]
    \centering
    \includegraphics[
        width=\textwidth,
        height=0.8\textheight,
        keepaspectratio
    ]{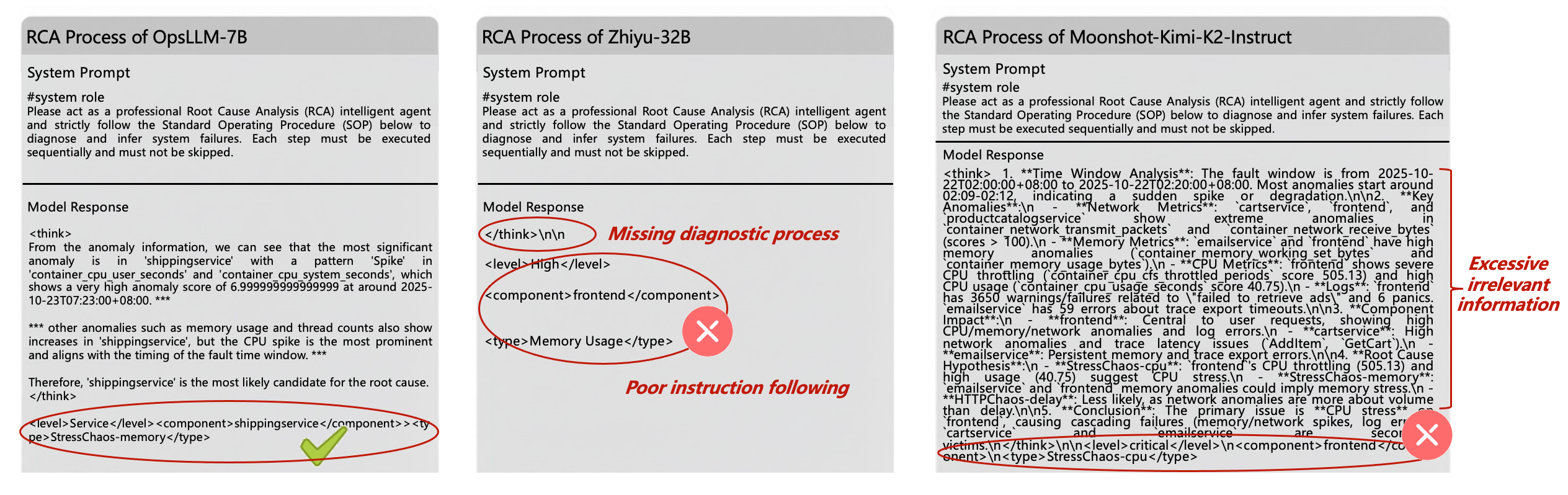}
    \caption{Comparison of practical RCA reasoning examples across different LLMs.}
    \label{rca}
\end{figure*}

\section{Results and Analysis}
\label{results}

\textit{Effectiveness Validation (RQ1).}
We compare OpsLLM with base, open-source, and closed-source LLMs on both QA and RCA tasks across Easy/Mid/Hard difficulty levels. The results are shown in Table \ref{tab:llm_acc_comparison}.

\textit{QA Task Improvement.}
As shown in Table \ref{tab:llm_acc_comparison}, OpsLLM variants show improvements over their base Qwen2.5-Instruct counterparts. Specifically, OpsLLM-7B achieves accuracies of 98.0\% (Easy), 97.6\% (Mid), and 66.3\% (Hard), representing gains of +1.2\%, +0.6\%, and +0.2\% over its base model, Qwen2.5-7B-Instruct (96.8\%, 97.0\%, 66.1\%), respectively; OpsLLM-14B scores 99.2\% (Easy), 98.4\% (Mid), and 79.0\% (Hard), outperforming Qwen2.5-14B-Instruct (97.6\%, 97.2\%, 73.3\%) by +1.6\%, +1.2\%, and +5.7\%, respectively; OpsLLM-32B attains the highest scores on the Easy (99.6\%) and Mid (99.8\%) subsets and achieves 88.7\% on the Hard subset. This translates to improvements of +1.6\%, +1.0\%, and +11.9\% over its base model, Qwen2.5-32B-Instruct. These improvements, particularly on the challenging Hard subset, validate the quality and breadth of our collected fine-tuning dataset, which enables effective generalization.

\textit{RCA Task Improvement.} The superiority of OpsLLM is even more pronounced on the reasoning-intensive RCA task. Our models achieve state-of-the-art performance on the Easy, Mid, and Hard subsets: OpsLLM-32B leads all models with remarkable accuracies of 77.4\% (Easy), 79.0\% (Mid), and 32.3\% (Hard), far exceeding its base model's performance of 11.3\%, 8.7\%, and 6.0\%; OpsLLM-14B follows closely, achieving 53.2\% (Easy), 56.0\% (Mid), and 21.1\% (Hard), a dramatic leap from its base model's 9.7\%, 4.3\%, and 4.7\%; OpsLLM-7B also shows significant gains, reaching 38.7\% (Easy), 41.3\% (Mid), and 12.9\% (Hard), compared to its base's 8.1\%, 4.0\%, and 4.4\%. This transformative improvement in reasoning capability is a direct result of our reinforcement learning training strategy, which explicitly compels the model to perform structured reasoning within \texttt{<think></think>} tags before generating the final answer. The critical importance of this explicit thinking process is corroborated by the performance of other leading models that employ similar mechanisms. For instance, on the Hard RCA subset, Qwen3-Next-80B-A3B-Thinking (27.4\%) and Qwen3-Max-2025-09-23 (22.2\%), both known for their reasoning capabilities, are among the top performers. This result strongly validates our design choice to prioritize and formalize the reasoning step within OpsLLM. For some LLMs in Table~\ref{tab:llm_acc_comparison}, the accuracy on RCA-Hard is higher than that on RCA-Mid. This is because the difficulty levels are defined based on the degree of distributional shift from our models' training data. As a result, the difficulty levels may not align well for other LLMs.

\subsection{Transferability Analysis (RQ2)}

\textbf{Transferability on  test sets.} As shown by the Hard-level QA~\cite{opseval} and RCA~\cite{aiops2025} tasks in Table~\ref{tab:llm_acc_comparison}, OpsLLM consistently outperforms the base LLM. Notably, the Hard-level evaluation is conducted on publicly available and highly challenging datasets that are not observed during training, indicating that OpsLLM can effectively transfer to unseen datasets. Furthermore, this improvement suggests that the fine-tuning dataset covers a broad range of software operations knowledge, enabling OpsLLM to answer a wider variety of operational questions. In addition, OpsLLM learns effective RCA methodologies, thereby avoiding cases of misaligned reasoning.

\textbf{Transferability on  model scales.} The 7B and 14B models are commonly adopted in existing LLMs and have been widely applied in practice~\cite{7b,14b}. Considering that some real-world software operations scenarios impose higher performance requirements on LLMs, we additionally train a 32B-parameter model. As shown in Table~\ref{tab:llm_acc_comparison}, compared with the base LLM, OpsLLM-32B achieves improvements of 1.6\%, 1\%, and 11.9\% on the three difficulty levels of QA tasks, and improvements of 66.1\%, 70.3\%, and 26.3\% on the three difficulty levels of RCA tasks. These results indicate that our workflow and training data remain effective for larger-scale LLMs. Moreover, benefiting from the stronger capability of the base LLM, OpsLLM-32B outperforms state-of-the-art closed-source LLMs.

\textbf{Transferability on  Different Microservice Systems.} We constructed 50 RCA cases on Train Ticket\footnote{\url{https://github.com/FudanSELab/train-ticket}} using Kubernetes-native fault injection across 32 microservice components and 5 fault types. For zero-shot evaluation, each prompt contained only observed anomaly information extracted from Kubernetes runtime state, event records, and service logs, along with candidate root-cause levels and components, fault-type definitions, and the required output format, without any Train Ticket examples. OpsLLM-14B achieved a 32\% accuracy (10\% improvement over the base model), surpassing all open-source and closed-source LLMs in Table \ref{tab:llm_acc_comparison}.

\vspace{-0.05in}

\subsection{Ablation Study (RQ3)}

\textbf{Contribution of Fine-tuning Data Components.} We remove each of the five subset from the fine-tuning dataset and fine-tuning Qwen2.5-14B-Instruct with LoRA~\cite{lora} based on LlamaFactory~\cite{llamafactory}. The trained models are evaluated on QA-Mid and RCA-Mid to quantify the contribution of each data component.

We have the following observations: (i) The Injection data (without data leakage) is essential for the RCA task; removing this subset during SFT reduces the RCA accuracy. (ii) We observe that incorporating textual data substantially improves accuracy on the QA task. This suggests that balancing the model's capabilities across different operational tasks requires of a well-proportioned dataset.

\textbf{Contribution of RL Design.} We first fine-tune Qwen2.5-14B-Instruct on the constructed fine-tuning dataset, and then conduct experiments on RCA-Mid. We make two observations: (i) When using the GRPO algorithm to simultaneously assign rewards to output format and final results, including both the root-cause component and fault type, the performance improvement is limited. This is because the model tends to converge to rewards obtained from partially correct and stable outputs, resulting in relatively low accuracy when both the component and fault type must be correct. (ii) Combining the GRPO algorithm with a multi-dimensional reward and DPRM helps the model capture key aspects of RCA problem-solving.

\textit{\textbf{Why This Design Suits the RCA Task.}} (i) RCA is not a single-step label prediction task, but a diagnostic process with explicit dependencies: the model must first determine the fault level, then localize the anomalous component at the corresponding granularity, and finally identify the fault type. A prediction is considered correct only when all these elements are correct simultaneously. Therefore, directly optimizing the full objective leads to an excessively large search space and overly sparse rewards in the early stage of training. The introduction of curriculum learning~\cite{curriculum} essentially organizes the training objective according to diagnostic difficulty, thereby reducing the optimization difficulty. (ii) The gating mechanism in each stage prevents reward propagation from deviating from the true diagnostic chain, thereby substantially improving credit assignment and preventing rewards from being incorrectly assigned to diagnostic behaviors that should not be encouraged. (iii) Final-result rewards alone cannot rule out cases where the model arrives at the correct answer by chance. The introduction of DPRM not only provides lightweight expert-style process evaluation, but also helps the model acquire more generalizable RCA capabilities. (iv) GRPO~\cite{grpo} updates the policy through within-group relative comparisons, allowing the model to more stably prefer outputs that are overall better across multiple reward dimensions. This makes it particularly suitable for progressive diagnostic tasks such as RCA.

\textbf{Contribution of Model Evaluation Components.} To validate the effectiveness of our MCQ-based evaluation, we conducted an ablation study using similarity-based metrics (BERTScore) on 100 operational QA pairs using the Qwen2.5-Instruct series. Under similarity-based evaluation, the performance ranking was 14B(0.74) $>$ 7B(0.70) $>$ 32B(0.68), contradicting the expected scaling law. In contrast, our MCQ evaluation (judged by exact option matching) produced a monotonic ranking of 32B(98.0\%) $>$ 14B(97.0\%) $>$ 7B(95.0\%), which aligns perfectly with the scaling law. This clear divergence confirms that similarity scores are unreliable for correctness, while our MCQ transformation provides an objective and accurate measure of LLMs' underlying knowledge and reasoning capability.

\begin{figure}[t]
    \centering
    \includegraphics[width=\linewidth]{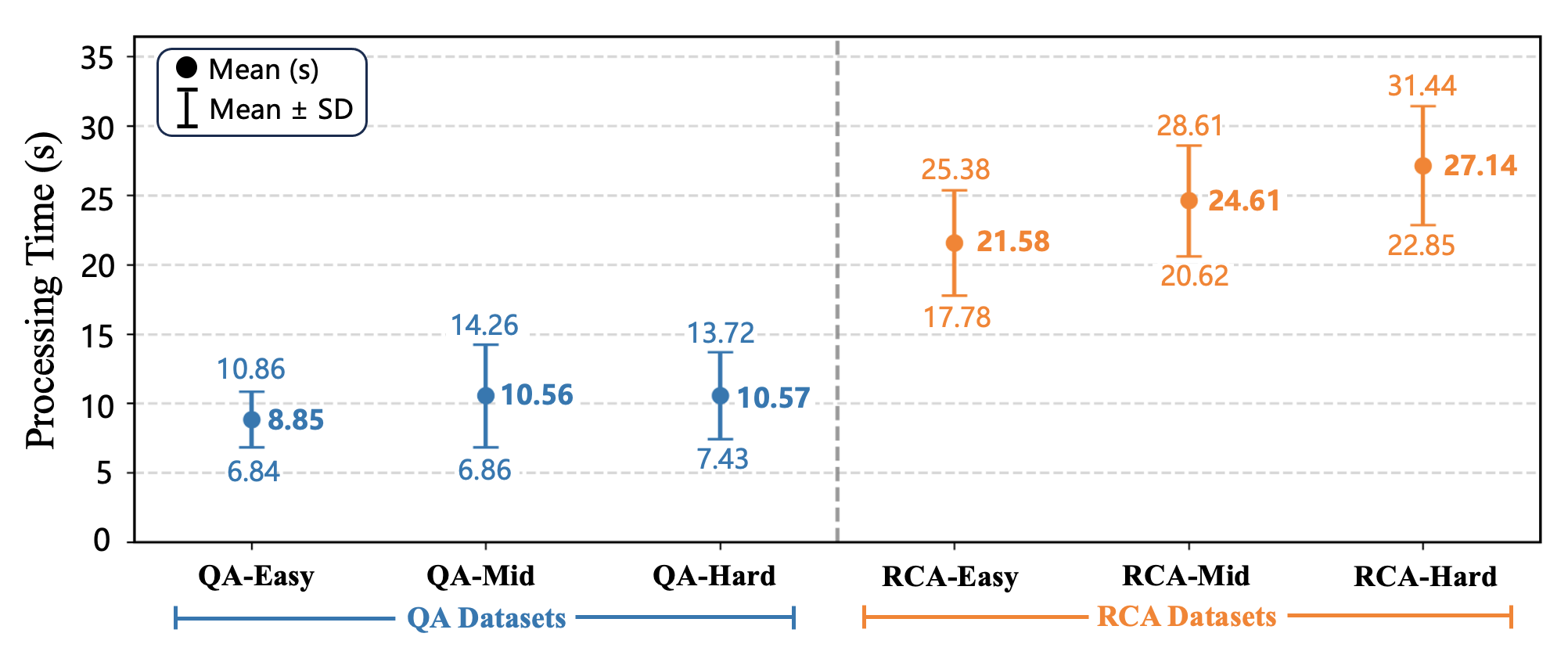}
    \vspace{-0.35in}
    \caption{ \textbf{Inference Latency Across QA and RCA Datasets of Different Difficulty Levels.} Error bars represent Mean ± Standard Deviation (SD), i.e., the range [Mean $-$ SD, Mean $+$ SD].}
    \label{performance.png}
\end{figure}

\begin{figure}[t]
    \centering
    \includegraphics[width=\linewidth]{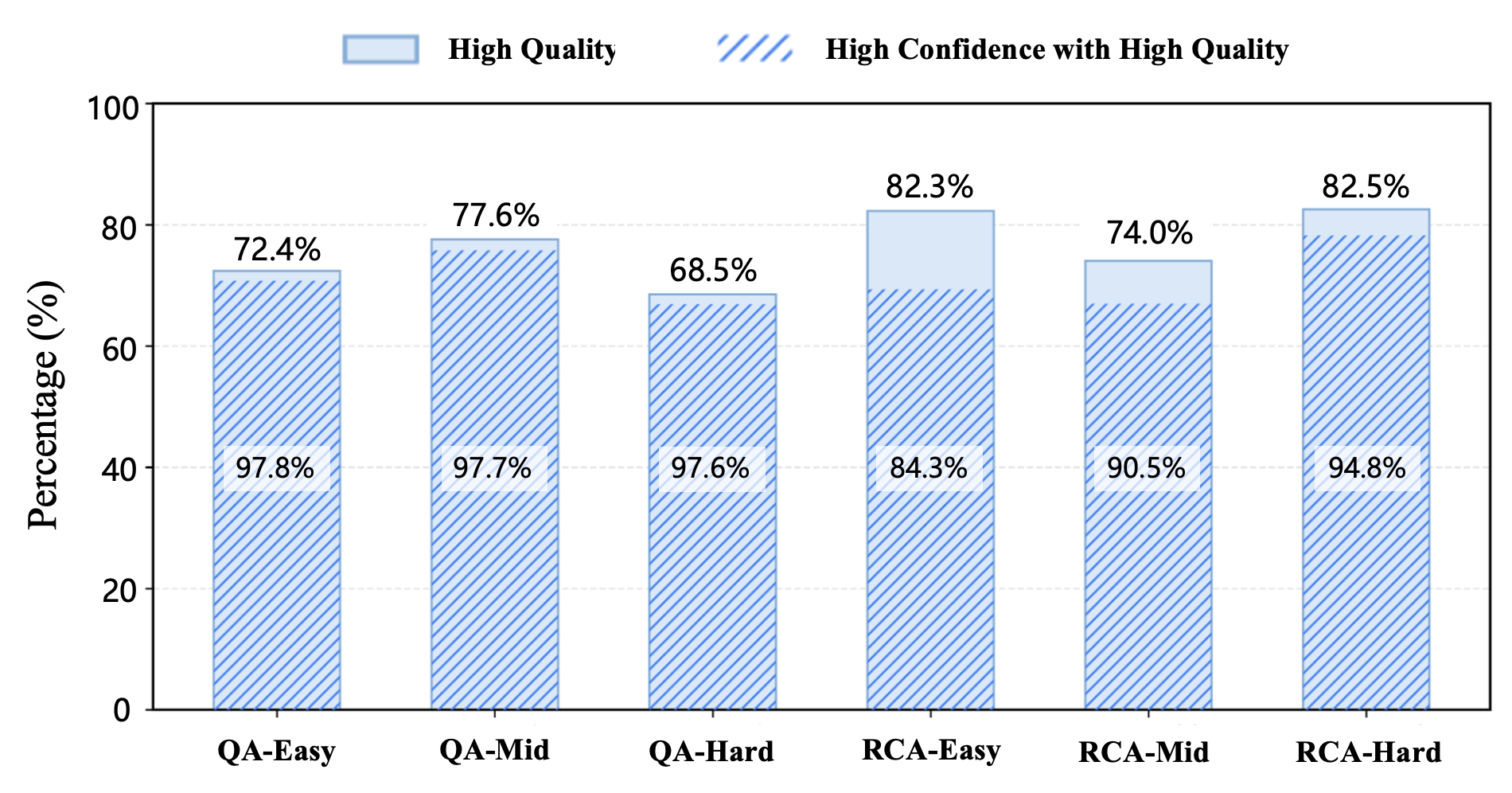}
    \vspace{-0.35in}
    \caption{Response quality assessment of OpsLLM-14B on six datasets using GLM-5.2 as an LLM judge. The solid bars denote the percentage of responses judged as high quality, and the hatched regions indicate the proportion of high-confidence judgments within the high-quality responses. (RCA-Hard samples exceeding the model context length were excluded.)}
    \label{quality.png}
\end{figure}

\subsection{Validation of DPRM}
\textbf{Experimental Setup.} We randomly sampled 200 reasoning chains from the DPRM training set with balanced coverage across the five score levels (0–5). Two human experts independently scored each chain using the same five rubrics employed for DPRM training.

\textbf{Results and Analysis.} The results demonstrate a strong alignment between DPRM's assessments and human experts. The primary metric, Spearman's rank correlation coefficient, between DPRM scores and the average human expert scores is 0.88 , which is significantly high. This indicates a near-monotonic relationship and confirms that DPRM's ranking of reasoning quality is highly consistent with human preferences.

\subsection{RCA Failure Analysis}
\label{rca_fail}

As shown in Fig.~\ref{rca}, we select fault injection data to compare OpsLLM-7B with two other LLMs. The results clearly show that OpsLLM-7B ensures both the correctness of the final outcome and the reliability of the reasoning process with a smaller model size.

\textit{\textbf{Why Current LLMs Fail on the RCA Task.}} (i) LLMs across different capability tiers tend to mistake symptoms for causes, imposing interpretations on the data that appear plausible but do not reflect the actual situation~\cite{whyfail}. In contrast, OpsLLM strengthens the ability to trace root causes back from anomalous phenomena. (ii) RCA is not a standard single-step QA task, but instead requires jointly analyzing operational data under complex dependency relationships. Existing LLMs are primarily trained through autoregressive prediction on general corpora and general instruction alignment, and therefore lack task-specific modeling for RCA. In contrast, the training process of OpsLLM decomposes challenging RCA capabilities into training signals that can be more effectively absorbed by the model. (iii) Existing LLMs are prone to the Limited Telemetry Coverage failure mode~\cite{whyfail}, where they rely only on local observations while ignoring other critical telemetry information. OpsLLM, by combining process supervision with stage-wise optimization, encourages the model to form a more complete evidence chain, thereby mitigating reliance on a single telemetry source.

\subsection{Evaluation of Practical Operational Capabilities}

\textbf{Practical Deployment Metrics.} We generated complete responses for each sample in the six datasets using vLLM based on OpsLLM-14B and recorded the inference latency of each inference task. The results are shown in Fig. \ref{performance.png}. QA tasks require on average around 8-11 seconds per sample, whereas RCA tasks incur higher latency (21–27 seconds on average) due to their substantially longer inputs and multi-step reasoning processes. In addition, we have released the reasoning traces of several questions in our source code.

\textbf{LLM-as-a-Judge Assessment.} Based on the responses generated by OpsLLM-14B on the six datasets, we employed GLM-5.2 as an LLM judge to evaluate the quality of each model response. Meanwhile, GLM-5.2 was also prompted to provide a confidence score for each judgment. As shown in Fig.~\ref{quality.png}, 72.2\% of all responses were assessed as high quality by GLM-5.2, among which 96.5\% were assigned high confidence. These results indicate that OpsLLM-14B not only achieves a high accuracy but also maintains a high response quality.

\section{Related Work}

Software operations are critical to the reliability of modern large-scale systems, directly impacting service quality, user experience, and business continuity. LLMs leverage their strengths in knowledge integration and reasoning to become an important tool for improving operational efficiency and enhancing system reliability.

\textbf{LLMs for Software Operations.} There are several open-source LLMs for the operations domain available in the community. Zhiyu2.0~\cite{zhiyu} builds an LLM-collaborative data pipeline, a configurable reward system, and an MCP-based tool environment, and adopts GRPO-based agentic RL training. DeepSeek-R1-Distill-SRE-Qwen-7B~\cite{SRE-7B} is fine-tuned on the OWL~\cite{owl} dataset. DeepSeek-R1-Distill-SRE-Qwen-32B-INT8~\cite{SRE-32B}, fine-tuned on the Devops\_LLM~\cite{devopsllm} dataset, targets enterprise system management and cloud-native operations platform development. In addition, camel-ai has released aiops-qwen-4b~\cite{aiops-qwen-4b}. However, these LLMs still exhibit limited performance on publicly available operational task benchmarks.

\textbf{Datasets for Software Operations.} Operational knowledge is important for building LLMs for the software operations domain and is widely distributed across different modalities, including observability data and textual data. In the open-source community, a number of textual operational datasets~\cite{devopsllm,aiops,aiops_camel,aiops_finetuning_qaset,smoltrace-aiops-tasks,opseval} have emerged, covering diverse operational scenarios such as incident response, automated operations, and cross-layer correlation analysis. Meanwhile, observability-oriented operational datasets, including ITBench-Trajectories~\cite{itbench-trajectories-2025} and OWL~\cite{owl}, focus on scenarios such as log parsing, root cause analysis evaluation, and security and compliance assessment. However, these datasets do not simultaneously cover both observability data and textual data, making them incomplete from the data modality dimension.

\section{Discussion}

\textbf{Overhead.} We collected a total of 65,753 samples from both observability data and textual data, which were filtered to obtain a final dataset of 15,027 samples. The construction of the entire dataset took approximately three months (including raw data collection, fault injection, data filtering, and the development of the data processing framework). We performed post-training of the base models on NVIDIA A100-80GB GPUs by first fine-tuning the base models with the constructed dataset using LoRA~\cite{lora}, followed by RL training based on the fine-tuned models.

\section{Conclusion}

In this work, we propose \textit{OpsLLM}, a domain-specialized LLM for software operations with core capabilities in QA and RCA. We also present the first end-to-end framework for constructing domain-specific LLMs tailored to software operations, which automates the entire workflow from data construction and model post-training to systematic evaluation. We release \textit{OpsLLM} models at three scales (7B, 14B, and 32B) to accommodate diverse performance requirements in real-world scenarios, along with a 15K fine-tuning dataset that spans multiple modalities, including observability data and textual data. Experimental results demonstrate the effectiveness and transferability of \textit{OpsLLM}.

\section{Acknowledgments}

This work was supported in part by
National Key Research and Development Program of China
(Grant Number: 2024YFB4505904), the National Natural Science Foundation of China under Grant 62272495 and the
Guangdong Basic and Applied Basic Research Foundation
under Grant 2023B1515020054. The corresponding author is Pengfei Chen.

\vspace{12pt}

\bibliographystyle{IEEEtran}
\bibliography{sample-base}

\end{document}